\documentclass{article}

\usepackage[utf8]{inputenc}
\usepackage{geometry}
\newgeometry{left=0.95in,right=0.95in,top=1.5cm,bottom=1.5cm,headheight=0.2cm}
\usepackage{graphicx,float}
\usepackage[ruled,vlined]{algorithm2e}
\usepackage{authblk}
\usepackage{mathtools}
\usepackage[colorlinks,
            linkcolor=green,
            anchorcolor=green,
            citecolor=green]{hyperref}

\begin{document}

\title {Standard and Event Cameras Fusion for Dense Mapping}
\author{Yan Dong}

\affil{Department of Automation, Tsinghua University, China}

\date{}
\maketitle

\begin{abstract}
Event cameras are a kind of bio-inspired sensors that generate data when the brightness changes, which are of low-latency and high dynamic range (HDR). However, due to the nature of the sparse event stream, event-based mapping can only obtain sparse or semi-dense edge 3D maps. By contrast, standard cameras provide complete frames. To leverage the complementarity of event-based and standard frame-based cameras, we propose a fusion strategy for dense mapping in this paper. We first generate an edge map from events, and then fill the map using frames to obtain the dense depth map. We propose "filling score" to evaluate the quality of filled results and show that our strategy can increase the number of existing semi-dense 3D map.

\hfill \break
\noindent \textbf{Keywords:} Multi-sensor Fusion, Event Camera, SLAM, Depth Estimation
\end{abstract}

\begin{figure}
  \centering
  \includegraphics[width=0.9\linewidth]{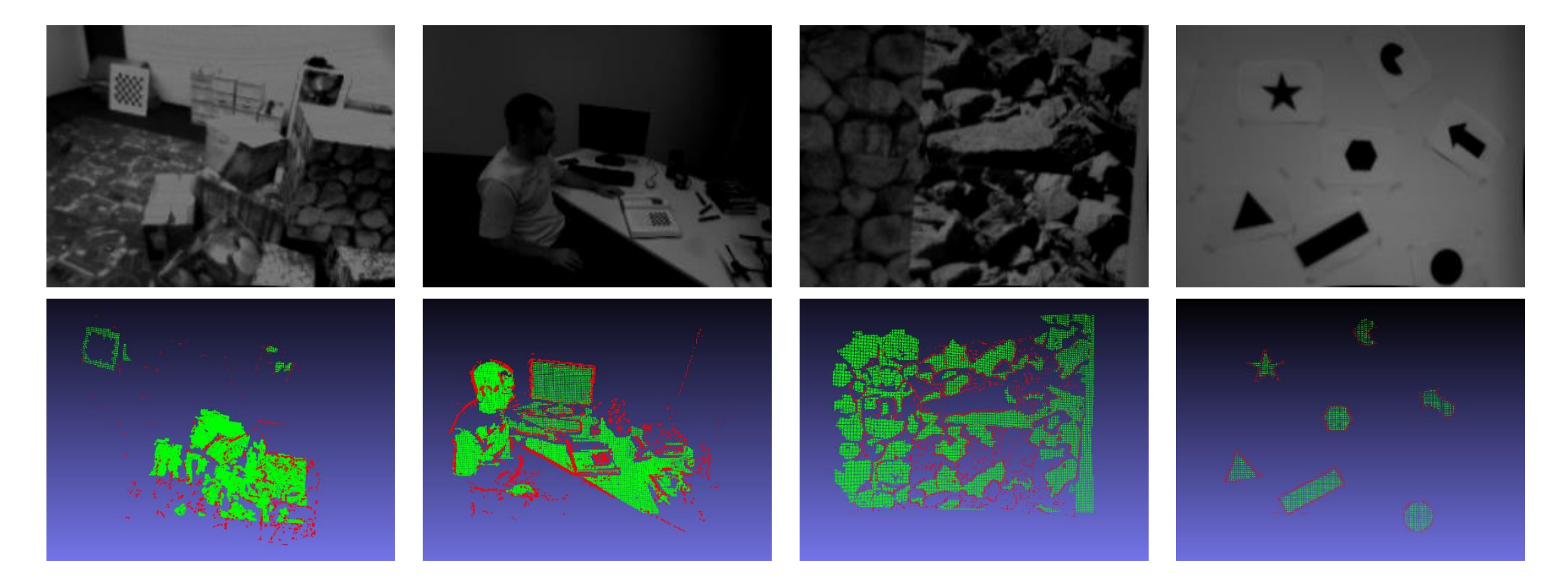}
\caption{Dense mapping on datasets. Top: original images. Bottom: dense map after fusing frames. Semi-dense map from the event camera is in red, and green points are added using frames.}
\label{fig:demo}
\end{figure}

\section{Introduction} \label{Sec1}

Mapping is a core block of many vision tasks such as simultaneous localization and mapping (SLAM). Frame-based mapping has been studied for over decades, but event-based mapping is gaining more and more attention recently. Event-based mapping can be divided into two categories, monocular or stereo. The monocular mapping methods include reconstruction-based mapping \cite{Reconstruct}, contrast maximization framework \cite{Framework, Focus}, and multiple view stereo ways \cite{EMVS}. However, maps from these methods are usually "edge map" which are sparse. A recent work \cite{Stereo} used optical flow to generate dense depth by performing dense disparity estimation and use events to track the disparity between frames. 

We notice that standard cameras provide frames containing all information about the scene, while event cameras only capture the changes which usually happen at edges. We hope to fuse two types of sensors to get the dense 3D map. The main contributions of this paper are: 1. we proposed a dense mapping method by using frames to fill the sparse 3D map from the events stream; 2. we introduced "filling score" to evaluate the quality of filling a semi-dense map. The mapping results are shown in Fig. \ref{fig:demo}.

\section{Method} \label{Sec2}

\subsection{Event Camera Mapping}
We choose EMVS \cite{EMVS} to obtain the sparse 3D map from events stream, but we point out that any event-based 3D mapping method can be used. We explain the EMVS method briefly here, and reader can refer to \cite{EMVS} for more details.

When the camera is moving, events generated in the event camera are caused by an "event source", a 3D map point in 3D scene, if not considering the noise. If we back project events caused by the same "event source", the rays must intersect at the 3D map point in the scene.

EMVS first chooses a reference view, and discretizes the volume containing the 3D scene and counts the number of viewing rays passing through each voxel. A voxel is determined to be a 3D point in the scene if it is a local maxima.

\subsection{Standard Frame Segmentation}
We select the frame captured at the reference view, and segment the image using the region growing segmentation method. Since events are only generated at pixels where brightness changes, it is natural to segment regions by their grayscale value. First each unlabeled pixel is set to be a new region, and then it begins to grow if the intensity at adjacent pixels are similar than a threshold. The region stops growing if the intensities of all adjacent pixels are much larger or smaller. After segmentation, there exist many small regions which usually locate between two regions due to the sharp edge. We ingore those small regions and only fill large ones.

\subsection{Map Filling}
We project all map points obtained from events onto frames, and call them "projected events". A region will be filled if there are enough projected events on its contour with limited projected events inside it.

To determine the number of projected events on its contour, we notice that due to the discretization error of the volume, some projected events are not exactly located at the contour. As a result, we calculate projected events in the 3-pixel ring of the region. When the number of projected events on region contour is larger than a threshold (\emph{e.g} 30\% of the contour length), while the number of projected events inside a region is smaller than a threshold (\emph{e.g} 5\% of the region size), the region can be filled.

A non-parameter method is adopted to fill each region. For each pixel inside the region, we calculate the Euclidean distances to each projected event in this region and estimate the depth of pixels by a weighted summation:
$$ d(p) = \lambda \sum_{k} w_k \cdot D(m_k), p\in A, p \notin M, m_k \in M $$
\noindent where $A$ denotes all pixels in the region, $M$ contains all pixels with depth information (projected events), $D(\cdot)$ are their depths. $w_k$ is the weight, and $\lambda$ is the normalized factor. The weights depend on the distances between projected events and pixels. We use inverse Euclidean distance as follow:

\begin{equation}
    w_k = \begin{cases}
        1, & dist(p, m_k) < 1   \\
        1 / dist(p, m_k), & otherwise       \nonumber
    \end{cases}
\end{equation}
where $dist(p,m_k)$ is the Euclidean distance between $p$ and $m_k$. Different weighting methods are studied in the experiment and it shows that the mapping precision is not sensitive to the weighing methods. 

\begin{figure}[h]
  \centering
  \includegraphics[width=0.9\linewidth]{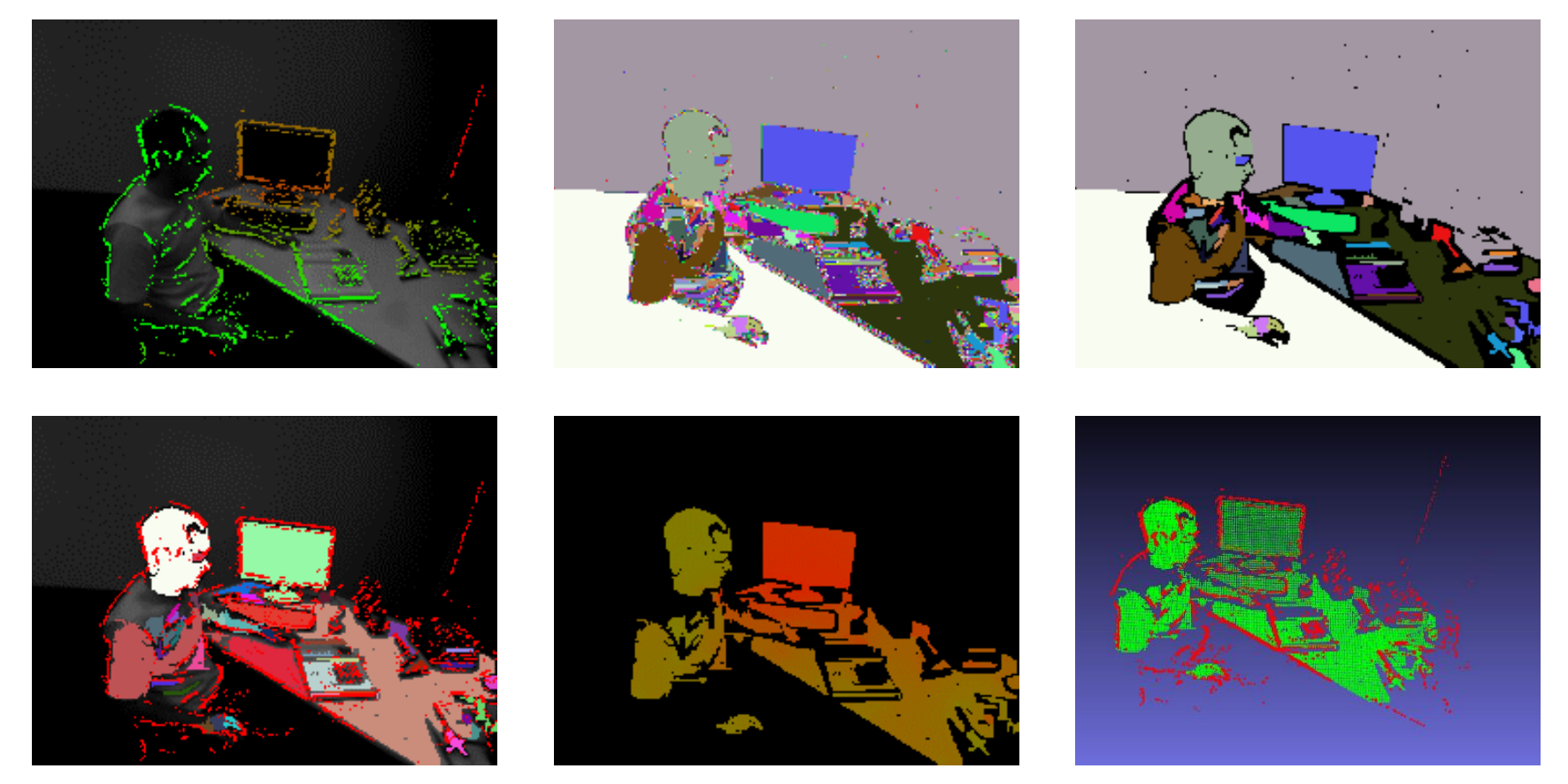}
\caption{The whole segmentation and filling process. Top: original frame with projected events by EMVS, raw segmentation results, valid regions (black pixels are invalid). Bottom: regions to be filled, the range image, the final point cloud.}
\label{fig:process}
\end{figure}

After obtaining the range image by map filling, a dense 3D map can be calculated easily. Figure \ref{fig:process} shows our segmentation and filling results.

\section{Experiment}

\subsection{Density Evaluation}
To evaluate the quality of the semi-dense map filling process, we define the "filling score" as follow:
$$ \beta = \frac{N_2-N_1}{Res+N_1 / Res} $$
where $N_1/N_2$ are the number of points in this 3D scene, $Res$ is the resolution of the image. 
 
Intuitively, more new points $(N_2-N_1)$ is better, but we should also consider the image resolution and the number of original points. It is more difficult to add the same number of 3D points if the original 3D map is less, that is what $N_1/Res$ indicates. The score is 0 if no new points are added. When $N_1=0 ,N_2=Res$, the score gets 1 but it is impossible to achieve a full mapping without any prior information. Thus $\beta \in [0, 1)$.

We test our method on Event Camera Dataset \cite{Datasets}. We use the released EMVS code \footnote{\url{https://github.com/uzh-rpg/emvs}} to obtain the edge 3D map. The results are shown in Table \ref{tab:density}.

\begin{table}[h]
    \centering
    \caption{Density Evaluation Experiment} \label{tab:density}
    \begin{tabular}{cccccc}
        \hline
        \centering{\quad sequence} & {\quad duration \quad} & {\quad $N_1$\quad } & {\quad $N_2$\quad} & {\quad $N_2/N_1$\quad} & {\quad $\beta$\quad} \\
        \hline
        boxes\_6dof          & 3.5s-5.5s  & 2289    & 10489   & 4.58        & 0.1898    \\
        dynamic\_6dof        & 6.2s-8.2s  & 2270    & 9125    & 4.01        & 0.1587    \\
        poster\_6dof         & 2.0s-4.0s  & 4748    & 19338   & 4.07        & 0.3377    \\
        shapes\_6dof         & 2.0s-4.0s  & 954     & 2137    & 2.24        & 0.0274    \\
        \hline
    \end{tabular}
    \label{table:dense}
\end{table}

From the table we can see that although "poster\_6dof", "dynamic\_6dof" and "boxes\_6dof" are all increased for 4 times, but the filling score are different. "poster\_6dof" has the largest score because the final 3D map occupies a larger region in image.

\subsection{Precision Evaluation}

To get the ground truth of 3D point cloud, we use "slider\_close", "slider\_hdr\_close", "slider\_far" and "slider\_hdr\_far" in the datatset. The event camera moves parallel to the poster with a fixed distance (23.1cm for "close" and 58.5cm for "far"). All sequences are about 6.5s and we segment each into 6 parts for every one second. For EMVS mapping, the grid range is set to be $240\times180\times100$, 15 for kernal size and 7 for the constant when finding the local maxima. The growing threshold is 3. The errors are shown in Table \ref{tab:error}.

\begin{table}[h]
    \centering      
    \caption{Absolute Error (cm)}        \label{tab:error}
    \begin{tabular}{p{3.5cm}  p{1cm}  p{1cm} p{1cm} p{1cm} p{1cm} p{1cm}}
        \hline
        \centering{sequence} & 0-1s  & 1-2s  & 2-3s  & 3-4s  & 4-5s  & 5-6s \\
        \hline
        slider\_close      & 1.67     & 1.25 & 1.62 & 1.57 & 0.68 & 1.01       \\
        slider\_hdr\_close & 1.34     & 1.59 & 2.28 & 0.54 & 1.06 & 1.58       \\
        slider\_far        & 1.79     & 1.24 & 1.23 & 1.88 & 1.30 & 1.37       \\
        slider\_hdr\_far   & 1.39     & 1.32 & 1.35 & 1.33 & 1.02 & 1.11        \\
        \hline
    \end{tabular}
\end{table}

We also studied different weighting methods: inverse distance, Gauss kernal and exponential kernal. Gauss kernal:
$w_k = \frac{1}{\sqrt{2\pi} \sigma} \exp(-\frac{dist(p, m_k)^2}{2\sigma^2})$ where $\sigma=5$. Exponential kernal: $w_k = \exp (-dist(p, m_k))$. Then the average errors of each sequence are shown in Table \ref{tab:weight_method}.
\begin{table}[h]
    \centering
    \caption{Mapping Error by Different Weighting Method (cm)}  \label{tab:weight_method}
    \begin{tabular}{ccccc}
        \hline
        \quad weighting method \quad  & \quad slider\_close\quad & \quad slider\_hdr\_close \quad & \quad slider\_far\quad & \quad slider\_hdr\_far \quad    \\
        \hline
        Inverse & 1.33          & 1.66               & 1.48        & 1.22             \\
        Gauss & 1.34          & 1.66               & 1.51        & 1.25             \\
        Exponential & 1.35          & 1.66               & 1.52        & 1.25             \\
        \hline
    \end{tabular}
\end{table}

The mapping precision depends largely on the semi-dense mapping, which is EMVS in this paper.

\section{Conclusion}
In this paper, we have introduced a novel approach for dense mapping by frames and events fusion. The semi-dense 3D map is obtained by EMVS method and the dense map is generated by fusing frames. Firstly, frames are segmented into different regions and depths at each pixel in regions are calculated by projected 3D map points. We also introduced the "filling score" to evaluate the filling process.

Our method focuses on obtaining a dense map using standard cameras, but not on event-based semi-dense mapping, which means any event-based mapping can be followed by this framework. However, there are some limitations of our filling strategy: 1. background regions cannot be perfectly filled due to the lack of information; 2. filling strategy is naive and the result is not real if the surface of region is complex.

The method can be improved in two ways. One is to adopt a different strategy, maybe learning based method, to fill regions. Another is that, this method only works on one reference view, which could be extended to using multiple views and merging local point cloud to get a full large one. 

We believe this paper is an important attempt to fuse two sensors for dense mapping and hope it can inspire other researchers.

\bibliographystyle{unsrt}
\bibliography{references.bib}

\end{document}